# Detect caterpillar, grasshopper, aphid and simulation program for neutralizing them by laser

Rakhmatulin Ildar, Ph.D, South Ural State University, Department of Power Plants Networks and Systems, Lenin prospect - 76, Chelyabinsk, Russia, 454080
ildar.o2010@yandex.ru

**Abstract.** The protection of crops from pests is relevant for any cultivated crop. But modern methods of pest control by pesticides carry many dangers for humans. Therefore, research into the development of safe and effective pest control methods is promising. This manuscript presents a new method of pest control. We used neural networks for pest detection and developed a powerful laser device (5 W) for their neutralization. In the manuscript methods of processing images with pests to extract the most useful feature are described in detail. Using the following pets as an example: aphids, grasshopper, cabbage caterpillar, we analyzed various neural network models and selected the optimal models and characteristics for each insect. In the paper the principle of operation of the developed laser device is described in detail. We created the program to search a pest in the video stream calculation of their coordinates and transmission data with coordinates to the device with the laser.

## 1. Introduction

Regardless of the type of pest (creeping, flying, jumping), all cultivars on all continents suffer from harmful insects. Mukandiwa et al. [1] and Davari et al. [2] described different types of pests and the scale of the harm that they do to the crop in detail. The most popular method of pest control is the use of pesticides. But pesticides carry a potential danger for human health and Azam et al. [3] considered this problem in detail. Many researchers involved in this field concluded that it is necessary to use a safe method of controlling harmful insects both for humans and for cultivated crops. Therefore, we set the task to develop an environmental device for pest control in the field. Directly to neutralize the pests, we decided to use a powerful laser, as it is environmentally friendly and waste-free. To detect pests, we decided to use neural networks, as in recent years they have established themselves as the most highly effective technologies in this field.

In the field of object detection today there are many researches in which various methods and models of neural networks are involved. But each new task requires a special approach since the effectiveness of machine vision depends not only on the size, type, speed of movement of the identified object, but also on the object background. We considered papers with a similar task for object detection according to our search criteria (pest in the field). Krishnan et al. [4] for pest control used detection algorithm realized by methods in MATLAB. This research used not typical image preprocessing process. Krishnan et al. created his own methods implemented in MATLAB. But the speed of image processing is not high and is not suitable for real work in the field. In contrast to the paper described above, Galphat et al. [5] used standard image processing functions of the OpenCV library, which allowed speeding up of the image processing process. But this research focuses only on the image processing process and the results of the search for pests in streaming video were not presented. Zhang et al. [6] presented a review for monitoring pests with remote sensing technology. But the solutions presented in this review can only give general information about the number of pests in the field. To assess the effectiveness of pest control, this method can be used in conjunction with the method presented in this manuscript. Cheng et al. [7] considered a similar task. But to implement the analysis of the total number of pests in the field, he proposed a new model of super-resolution, based on a deeply recursive residual network, due to which the image is restored in a higher quality. Qin et al. [8] detected the edges of spectral residual significance for pest's reorganization. This method allows you to obtain a recognition result with high accuracy. The object detection accuracy is very much dependent on the background on which the object is located. Nouri et al. [9], Roldán-Serrato et al. [10], Basati et al. [11], and Djekic et al. [12] applied classification algorithms to identify pests in images quite accurately. These researches considered a specific type of pest against a specific type of crop. The researchers used images in very high resolution, which requires spot focusing of the camera, and therefore this method is not easy to use in a field.

A paper of Cheng et al. [13] is closer to our manuscript. He used the deep neural networks to classify images of 10 different pests. But the used models of neural networks will not allow to use the results of this work for real-time tasks. Wang et al. [14] introduced algorithms for processing images in real-time to speed up the work of deep neural networks using the example of pest search. In the future, with an increase in the number of monitored objects per unit time, it is advisable to use this algorithm for our device. Some of the authors researched artificial conditions or greenhouses, in result of which videos or photos were originally obtained for analysis in a good quality [15].

The next manuscripts are as close as possible to the subject matter of our research. Jahangir et al. [16] used neural networks to detect mosquito in real-time. The author considered only one method for detecting the absolute difference estimate - and the results on determining the position-Z were not presented. This will not allow us to calculate the real position of the object in practice. Mathiassen et al. [17] considered the effect of the laser on the weed. This paper interested us from the point of view of the laser and its effect on the crop. During the pest neutralization, the crop should not be affected. As a result, we selected the optimal power and wavelength for the laser. Mullen et al. [24] developed a laser plant for controlling volatile pests. But the installation was large, expensive, and was tested in ideal conditions where the object was moving against a dark background. There is no possibility of use in real conditions. Hatton et al. [25] did similar research, but unlike the work above, he used low-cost tools to implement a laser pest control device. But in this research the same drawback is that the identification took place in ideal conditions, where the object was placed on a one-color background.

Worth adding that the Bill & Melinda Gates Foundation sponsored the research on the introduction of laser mosquito-neutralization technology as a promising method for malaria control. As a result, Photonic Sentry was created (https://photonicsentry.com/). For several years company showed a demo video with mosquito neutralization without describing the installation. But we didn't find scientific articles about this installation.

It can be concluded that there is no optimal method of the pest recognition task in the field. Each author solves this problem based on existing conditions. All this reduces the value of the results obtained since they cannot be applied to another type of pest. As a result, the purpose of this manuscript develops the low-cost device which identifies pests in the video stream in the amount of 10 pieces per second, calculates the coordinates of these pests, and transmits the coordinate data to the laser device for next neutralization.

For research we selected some of the most popular pests: cabbage caterpillar, aphid, and grasshopper. Firake et al. [18] and Ibrahim et al. [19] described the negative consequences of cultivated crops from cabbage caterpillar in detail. Wei et al. [20] and Barberà et al. [21] researched aphid and described the effect of this pest on the yield for various crops Sharma et al. [22] and Bradshaw et al. [23] considered grasshopper and its effect on the crop. All authors of these papers concluded that the damage from pests for crops will only increase since pests are becoming more resistant to modern methods of controlling.

The main part of the manuscript is divided into two parts. The first part is the pest recognition in the video stream - "Pest detection" and the second part is its neutralization recognition pest by the developed laser installation - "Pest neutralization by laser device".

## 2. Machine vision for pest detection task
### 2.1. Processing pest images for feature extraction

For pest detection we used images received from real-time video with a resolution of 400 by 400 pixels. For correct image recognition, it is initially necessary to distinguish the characteristic features of each pest against the background of a cultivated plant. We considered the following functions of OpenCV library which is the most popular in works on the search for pests on images (https://github.com/Ildaron/Pest_OpenCV-image-preprocessing-python):

**- Detection by color:**

- Color filters (cv2.createTrackbar);
- Color detection (cv2.inRange, cv2.findContours).

- **Smoothing Images**:

- Convolution (cv2.filter2D);
- Image Blurring (cv2.blur);
- Bilateral Filtering (cv2.bilateralFilter);
- Laplacian (cv2.Laplacian).

- **Morphological Transformations:**

- Erosion (cv2.erode);
- Dilation (cv2.dilate);
- Opening (cv2.MORPH_OPEN);
- Morphological Gradient (cv2.MORPH_GRADIENT);
- Top Hat (cv2.MORPH_OPEN);
- Black Hat (cv2.MORPH_BLACKHAT).

**Edge Detection:**

- Contours (cv.drawContours);
- Figure (cv2.rectangle, cv2.minEnclosingCircle, cv2.circle);
- Fill (cv2.floodFill).

**Thresholding:**

- Binary (cv2.THRESH_BINARY);
- Binary_inv (cv2.THRESH_BINARY_INV);
- Trunc (cv2.THRESH_TRUNC);
- Tozero_inv (cv2.THRESH_TOZERO_INV).

**Geometrical image transformation:**

- Resize (cv2.resize);
- Translation cv2.warpAffine();
- Affine Transformation (cv2.getAffineTransform);
- Perspective Transformation (cv2.getPerspectiveTransform).

**Image Derivatives:**

- Sobel derivatives (cv2.Sobel);
- Scharr derivatives (cv2.Scharr).

For example, in figure 1-the image processing with cabbage caterpillar with a thresholding filter is shown.

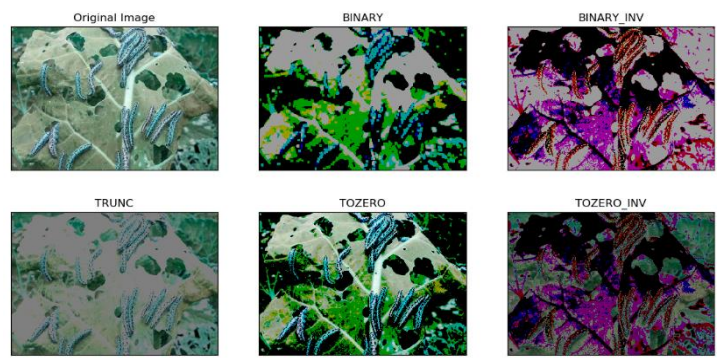

Fig.1. Image processing with cabbage caterpillar with the thresholding filter

To process the image with cabbage caterpillar we selected the next filters presented in figure 2.

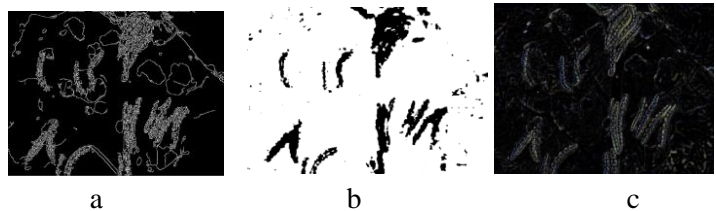

a b c

Fig.2. Image processing for cabbage caterpillar with the edge detection (cv2.Canny) – a, with the color filters (cv2.createTrackbar) – b, with the morphological transformations (cv2.MORPH_BLACKHAT) – c

**2.2. Neural networks models for pest detection**

After the image preprocessing process, we considered different types of convolutional neural networks (CNN) for the task of tracking pests. Today these networks have already become the standard in the field of machine vision and there are many different neural networks implemented in this technology. But not all networks are suitable for real-time tracking tasks. The use of convolutional neural networks on RaspberriePi is complicated due to the limited amount of RAM on the Raspberry Pi (1–4 GB) and the low processor frequency of 1.5 GHz. For example, the following popular models have the following size: ResNet50> 100 MB, VGGNet> 550 MB, AlexNet> 200 MB, GoogLeNet> 30 MB. Real-time detection with R-CNN, Fast R-CNN, Faster R-CNN, RetinaNet has the same recognition speed problem. The solution may be to use - NVIDIA Jetson TX1 and TX2 - a special platform for computing neural networks. The main disadvantage is that these devices cost a lot. Depending on the configuration, prices for the Jetson TX1 start at $ 300. The cost of the Raspberry PI 4 is around $ 40. Raspberry PI analogs - Orange PI and Banana PI have a wider line of unicameral computers of various types at a lower price but their characteristics are not higher. Therefore, we focused on the methods that can be implemented on the Raspberry Pi, which allows you to create an economical and compact device. We used a deep neural network, YoloV3, to detect pests. This neural network we used sequential $3 \times 3$ and $1 \times 1$ convolutional layers, as recommended by the authors of this neural network [26]. Diagram of the CNN in figure 3 is shown.

| | Type | Filters | Size | Output |
|---|---|---|---|---|
| | Convolutional | 32 | 3 × 3 | 256 × 256 |
| | Convolutional | 64 | 3 × 3 / 2 | 128 × 128 |
| | Convolutional | 32 | 1 × 1 | |
| 1× | Convolutional | 64 | 3 × 3 | |
| | Residual | | | 128 × 128 |
| | Convolutional | 128 | 3 × 3 / 2 | 64 × 64 |
| | Convolutional | 64 | 1 × 1 | |
| 2× | Convolutional | 128 | 3 × 3 | |
| | Residual | | | 64 × 64 |
| | Convolutional | 256 | 3 × 3 / 2 | 32 × 32 |
| | Convolutional | 128 | 1 × 1 | |
| 8× | Convolutional | 256 | 3 × 3 | |
| | Residual | | | 32 × 32 |
| | Convolutional | 512 | 3 × 3 / 2 | 16 × 16 |
| | Convolutional | 256 | 1 × 1 | |
| 8× | Convolutional | 512 | 3 × 3 | |
| | Residual | | | 16 × 16 |
| | Convolutional | 1024 | 3 × 3 / 2 | 8 × 8 |
| | Convolutional | 512 | 1 × 1 | |
| 4× | Convolutional | 1024 | 3 × 3 | |
| | Residual | | | 8 × 8 |
| | Avgpool | | Global | |
| | Connected | | 1000 | |
| | Softmax | | | |

Fig.3. Diagram of CNN YoloV3

The architecture of this neural network makes it possible to track objects in real-time. In our case, due to the low performance of the raspberry pi3, we focused on a simpler task - monitoring about 10 objects every second. We trained the model without using any initial weights in Google Collab. At the initial stage, it took time, about 16 hours, and required many images, we used 2800 images for every pest. After that we can use the obtained model weights to train quickly the neural model to detect a new type of pest. To do this, we need only to freeze the model with the initial weights and activate only the last layers of the model and train the last layers already for a specific type of pest. As a result, this will reduce the training time for recognizing a new type of insect. The complete code of the neural model (training the model, using the model) is implemented in python https://github.com/Ildaron/Pest_YoloV3_python_custom-data.

**2.3. Test neural networks models for pest detection**

At the initial stage, to check the operation of the neural network and configure the executive program on RaspberryPI, we created the program for simulation of the laser operation. The program implemented in Python3.6 language received coordinates from the neural network and then, as a simulation of the laser operation with a delay of 50 milliseconds, the position of the pest in the video was determined by drawing a red point. If the coordinate received from the python of the program coincided with the coordinates received from the neural network, it meant pest had been destroyed, figure 4.

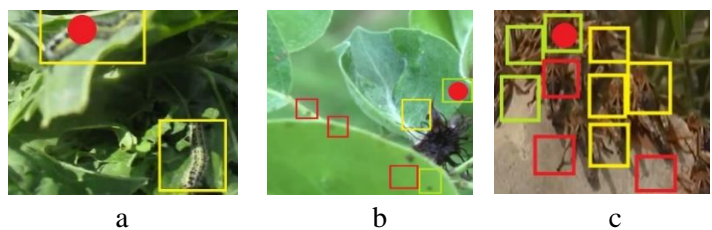

a          b          c

Fig.4. Program for pest detection: a - cabbage caterpillar, b - aphid, c - grasshopper. The yellow square - the pest is detected, the green square - the pest is neutralized, and the red square - the pest was not neutralized

We checked the trained neural network and the laser simulation on the video with pests. The results are presented in Table 1.

Table 1. Test results of the neural network and the laser simulation

| № | Average time to detect, ms | Pests amount | Total recognition (correct and not) | Recognition accuracy | Neutralized | Millisecond neutralization, ms |
|---|---|---|---|---|---|---|
| Cabbage caterpillar | 300 | 100 | 90 | 80 | 75 | 50 |
| Aphid | 350 | 120 | 85 | 45 | 30 | 50 |
| Grasshopper | 250 | 80 | 20 | 30 | 20 | 60 |

The higher result we achieved for the cabbage caterpillar: what happened due to the low speed of the caterpillars, their large size, and color contrast with the background. Aphid recognition accuracy is 2 times lower due to the lack of pronounced color contrast. Grasshopper moves too fast, because of this the camera and raspberries cannot determine their position in the dynamics. Later, for experimental research of the laser, we used only cabbage caterpillar.

**3. Development of a laser device for pest control**

To neutralize the pest, we decided to use a powerful laser with a wavelength of 450 nm and power 5 W. This power is enough to destroy the pest with a mass fraction of not more than 2 grams in no more than 25 milliseconds. The laser is fully environmentally friendly, energy-efficient. For development of the laser device we used the next tools, table.2

Table.2. Tools used in the development of the laser system

| 1 | Raspberry Pi 3 Model B+, Broadcom BCM2837B0 with a 64-bit quad-core processor (ARM Cortex-A53) with a frequency of 1.4 GHz – for control laser and for neural networks operation |
|---|---|
| 2 | Pi camera, Sony IMX219 Exmor – for pest detection |
| 3 | Galvanometer (speed - 20 kpps), the company – Unbranded/Generic, style DMX Stage Ligh type – DHR 814494 in protocol for laser control - ILDA DB25 – for laser control |
| 4 | Arduino Leonardo ATmega32u4 - to move the laser horizontally |
| 5 | 5 W laser (Jilin , MBL-W-457) with a wavelength of 450 nm. for pest neutralization |
| 6 | Step motor VerticalKit-600mm – for horizontal moving |
| 7 | Program language - python 3.6, vision library OpenCV 3.4.1 – for write control software and neural networks |

One of the difficulties in recognizing an object in space is calculating the z coordinate. To determine the correct distance to the object we used OpenCV stereo vision functions. For that two cameras are needed, and the method of mirror image separation are needed. The layout of the cameras is shown in figure 5.

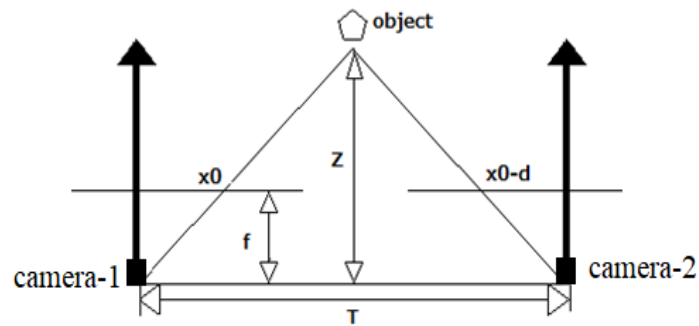

Fig.5. Positions of the cameras to determine the distance to the object: d - a value called disparity/offset, T - the distance between cameras, Z - distance to the object, f - focal length camera ratio

The distance between the cameras calculated by the following formula:
$$(T-d)/(Z-f) = T/Z \qquad (1)$$
The transformation of three-dimensional coordinates to the target in two-dimensional coordinates is based on direct kinematics. We developed a system for instant laser guidance based on a galvanometer, figure 6.

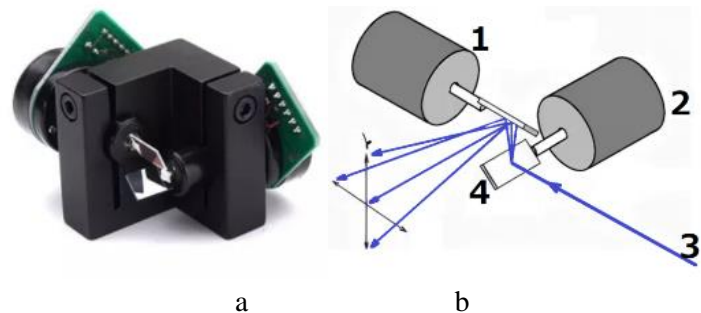

a  b

Fig.6. System for instant laser guidance: a - photo of the galvanometer, b - principle of operation, where 1 and 2 - DC motors, 3 - the laser beam, 4 - mirrors for laser reflection

The control of the positions of the galvanometer mirrors are described by the following expression:

$$P_{i+1} = P_i + \frac{d_i * n_{i+1} * (C_{i+1} - P_i)}{d_i * n_{i+1}}, \qquad (2)$$

where C is the three-dimensional coordinates of a point on the mirrors, n is the normal vectors of the units of the mirrors, and P is the target point on each mirror. Before using the formula, we made a calibration for the mirrors to work correctly, using the potentiometer in the laser board. The photograph of the developed device and operation scheme are presented in figure 7.

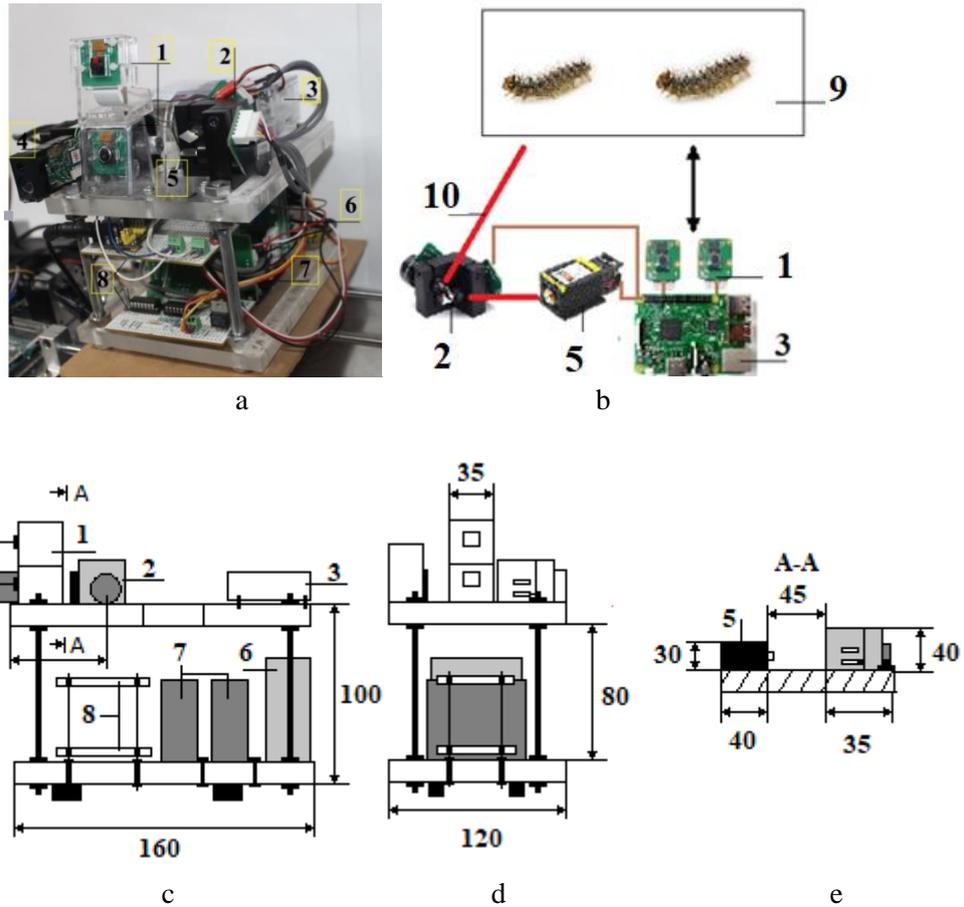

Fig.7. The developed device, where a – photo of device, b - structural scheme, c,d,e - device dimensions in mm: 1 - pi cameras, 2 - galvanometer, 3 - Raspberry Pi, 4 - laser rangefinder for checking the distance, 5 - laser device, 6 - power supply, 7 - galvanometer driver boards, 8 - analog conversion boards, 9 - object detection, 10 - laser beam

The single-board computer RaspberryPi processes the signal from the camera-1 and determines the position of the harmful insect-9 in the coordinates x, y, z and transmits a digital signal to the analog board - 8. This board has a digital-to-analog converter that converts the signal in the range 0-5V, after which the next board with the operational amplifier converts to a signal of +5V and -5V. This signal controls the galvanometer mirror -2 by the driver motor boards- 7. The galvanometer by mirrors changes laser beam direction - 10. The system is powered from the power supply - 6.

The operation principle of the developed device is described in figure 8. Before using the galvanometer, it should be calibrated by the potentiometer in the driver board before applying (according to the manufacturer's instructions in the galvanometer manual).

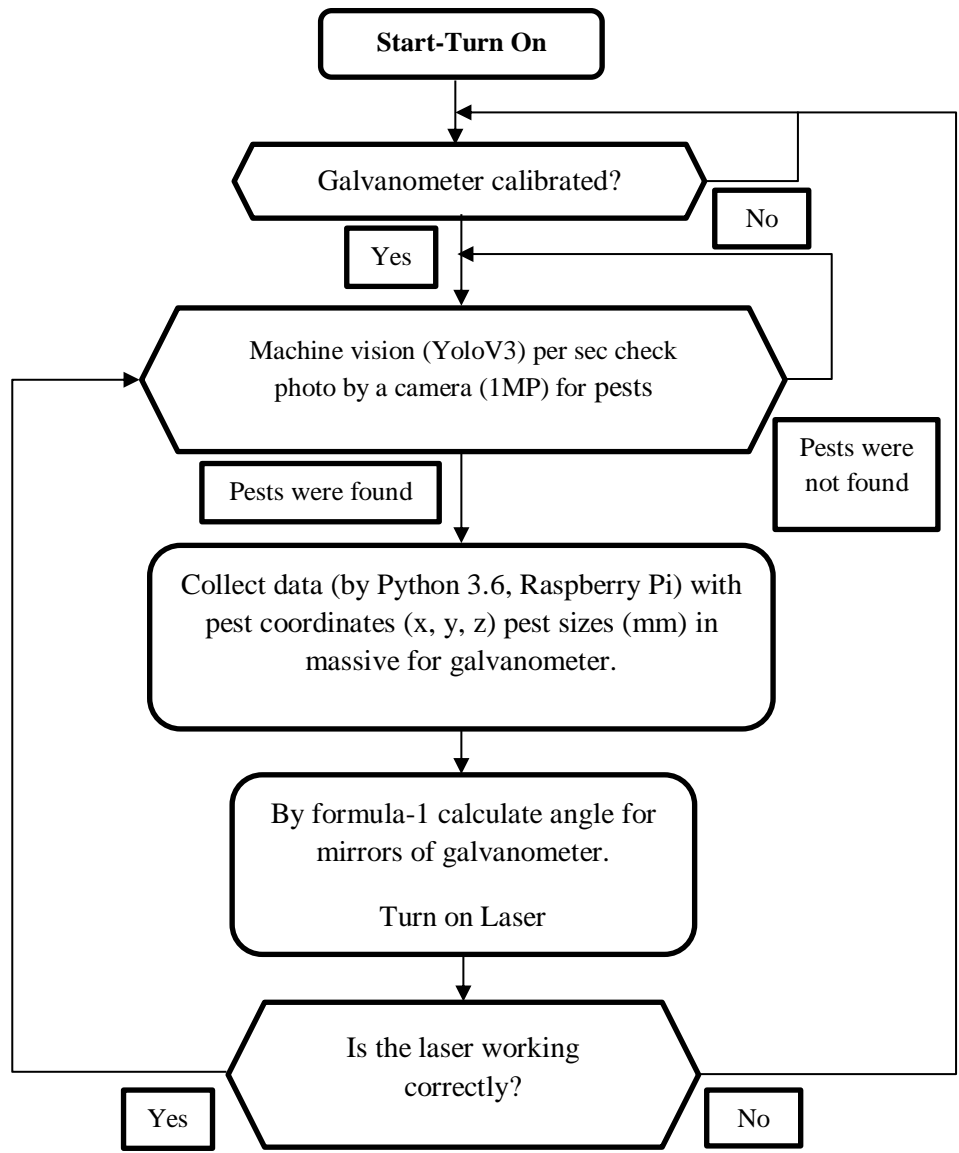

Fig.8. Algorithm of the laser device

**4. Experimental research of the laser device for pest control**

Cabbage caterpillar speed is not more than 3 mm per second. The caterpillar's speed can be neglected since the system we developed monitors an object 3 times per second and can to recognize about 10 objects at a time. From the moment of detecting the object and the laser operation about 30 milliseconds passes. For us, the most important point is the ability to neutralize a pest on the field. To be able to eliminate the field, this laser device should move. For implementation this function we researched the next moments:

- dependence of pest neutralization efficiency on the distance between the laser and the crop;
- dependence of pest neutralization efficiency on the speed of movement of the laser system.

Since we neglected the speed of cabbage caterpillar, we used an imitation of caterpillar figures as objects for detection and neutralization, which are in a random order for each experiment on the cultivated crops, figure 9.

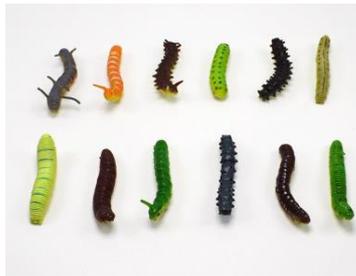

Fig.9. Example of objects for imitation of caterpillar figures

To research the effect of distance on the effectiveness of pest neutralization, we set the laser from the object at the following distances - 0.5 m, 1 m, 1.5m, 2m, 3, m, 5 m, 10 m. For each distance, we carried research 25 times, in which the pest each time is arranged in random order in a new place. We adjusted the laser spot focus manually - 2 mm in diameter. In the future, it is planned to use automatic laser focusing. Figure 10 shows the results of the research, for each distance, the result was averaged out of 25 attempts.

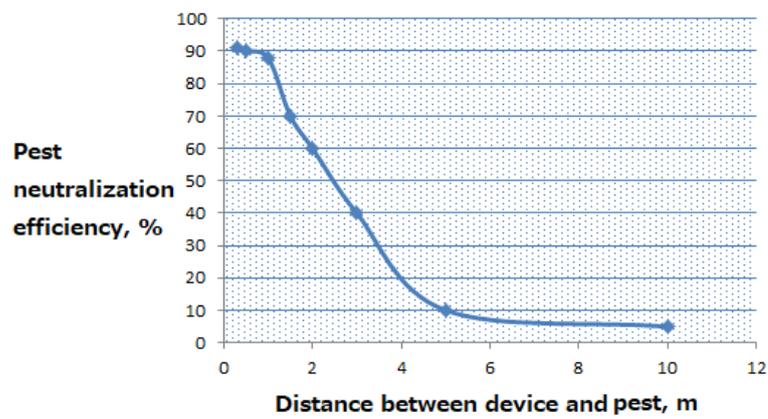

Fig.10. Dependence of the effectiveness of pest neutralization on the distance between the laser system and the pest

As expected, the accuracy of the camera decreases when the distance between the laser and the object increases. But it is important to find the optimal distance at which we can control the pest with high efficiency. In future practice, it is advisable to use cameras with autofocus, which will focus on the cultivated crop and only after that search for pests on the resulting image. At this research we used a camera without autofocus and therefore the accuracy of detection of pests decreases.

To bring the tests as close to natural conditions as possible for the experiment, we made a mechanism for the horizontal movement of the laser. This mechanism moves the laser between three points located at 300 mm from each other. Images and operation scheme of which are shown in figure 11.

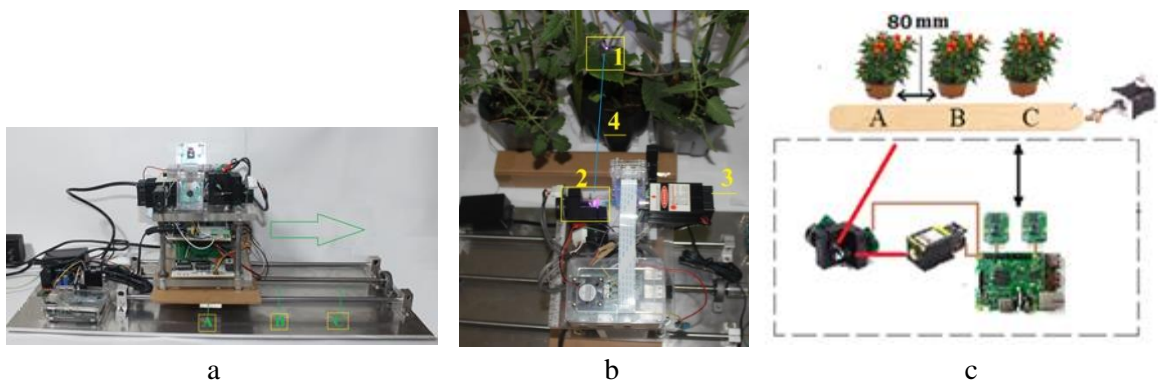

a          b          c

Fig.11. Images and operation scheme of laser device: a - front view, b -back view, c - operation scheme

For practical use on the field, an important point is the use of this development in a vehicle or unmanned aerial vehicle. We chose the appropriate distance to the weed in the previous experiment of 1 meter, which will subsequently allow the use of this device as a part of a vehicle or unmanned aerial vehicle. It was for this distance that we conducted studies on the effect of speed on the number of neutralized pests. Similarly, to the previous experiment, the caterpillars were randomly arranged on a tomato. For each distance, the experiment was carried out 30 times. Pest neutralization confirmed visually after each attempt.

Using the RaspberryPi program by Pulse-width modulation (PWM) we changed the motor speed of movement. We set the task to make the laser system manage to neutralize about 10 pests during movements. Graphically, this dependence will look as follows. The result obtained in the experiment is presented in figure 12.

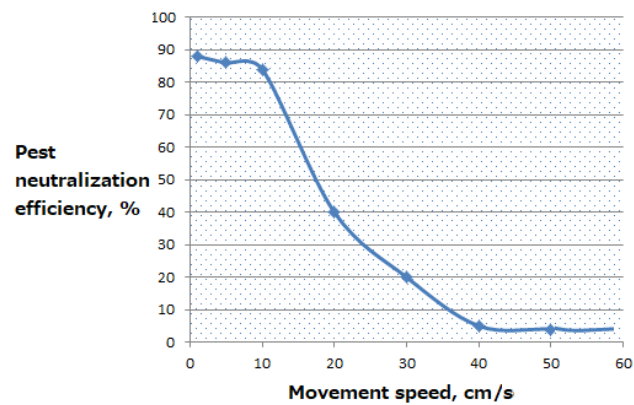

Fig.12. Dependence of the effectiveness of pest neutralization from the speed of movement of the laser system

The increase in speed directly affects the number of neutralized pests. To destroy the target, we fixed the laser beam on one object - 50 ms. At the same time, mirrors can change position 20000 times per sec, and a neural network can detect about 30 objects per sec, respectively, the number of neutralized objects depends on the laser power. But at the same time, a 15 W laser burns through both the pest and the useful plant, so it was decided not to use it. In this task, 5 W is the optimum power for the laser. At a speed of 10 centimeters per second to a distance 1 meter we managed to achieve the result of neutralizing pests in the amount of 10 units per second.

5. Discussion and conclusions

The results obtained in the research allow us to conclude that it is advisable to use the powerful laser device for combating pests. The main difference of this work is that the experiments were carried out in conditions close to natural. The background for pests was colorful. This method can help improve pest control in agriculture and contribute to the conservation of ecology and the environment. The proposed method is a new direction in the field of eliminating harmful insects in the fields of agriculture and can be considered as a low-cost option for pest control. The total cost of the equipment involved does not exceed 200 dollars. We analyzed various models of deep networks and optimal models for each insect. As a result, we were able to use a deep neural network on one board Raspberry PI computer. The combination of this neural network and analog circuitry allowed us to achieve a result in which this installation can be used in a natural environment with high efficiency. As a pest, we considered the most popular pests. The use of other types of insects requires only retraining of the last weights of the neural network. We obtained the value for the optimal distance between the laser installation and pests. In the manuscript it is shown that the laser device can operate in movement, which will allow it to be applied in the field- real conditions.

The next step in the research is planned to experiment directly on the field with a cultivated crop. The developed device will be installed on the vehicle on the field. To do this, we need to update the installation and run it in a protected IP54 enclosure. It is planned to replace Raspberry pi with STM32 microcontrollers. STM32 has X-CUBE-AI - AI - an expansion pack for STM32CubeMX. This extension can work with various deep learning environments such as Caffe, Keras, TensorFlow, Caffe. Thanks to this, the neural network can be trained on a desktop computer with the possibility of computing on the GPU. After integration, the optimized library for the 32-bit STM32 microcontroller can be used. Moreover, the use of a telephoto lens and a servomotor to move along the z-axis as part of the laser system can significantly increase the area of insect control.